\pdfoutput=1

\documentclass[twoside,11pt]{article}

\usepackage{jmlr2e}
\usepackage{amsmath}
\usepackage{url}

\newtheorem{thm}{Theorem}[section]

\newtheorem{lem}[thm]{Lemma}

\newcommand{\R}{\mathbb{R}}
\newcommand{\norm}[1]{\big\Vert#1\big\Vert}
\newcommand{\eps}{\varepsilon}

\newcommand{\argmin}{\operatornamewithlimits{argmin}}
\newcommand{\one}{\textbf{1}}
\newcommand{\Xt}{\widetilde{X}_i}
\newcommand{\Tt}{\widetilde{T}}
\newcommand{\Lt}{\widetilde{L}}
\newcommand{\Ut}{\widetilde{U}}
\newcommand{\wt}{\tilde{w}_i}

\newcommand{\Uh}{\widehat{U}}
\newcommand{\li}{\lambda_1^i}

\ShortHeadings{}{} \firstpageno{1}

\begin{document}

\title{LLE with low-dimensional neighborhood
representation}

\author{\name Yair Goldberg \email yairgo@mail.huji.ac.il \\
\addr Department of Statistics\\
The Hebrew University, 91905 Jerusalem, Israel\\
\AND
        \name Ya'acov Ritov \email yaacov.ritov@huji.ac.il \\
       \addr Department of Statistics\\
       The Hebrew University, 91905 Jerusalem, Israel\\
}

\editor{??}

\maketitle
\begin{abstract}
The local linear embedding algorithm (LLE) is a non-linear
dimension-reducing technique, widely used due to its computational
simplicity and intuitive approach. LLE first linearly reconstructs
each input point from its nearest neighbors and then preserves these
neighborhood relations in the low-dimensional embedding. We show
that the reconstruction weights computed by LLE capture the
\emph{high}-dimensional structure of the neighborhoods, and not the
\emph{low}-dimensional manifold structure. Consequently, the weight
vectors are highly sensitive to noise. Moreover, this causes LLE to
converge to a \emph{linear} projection of the input, as opposed to
its \emph{non-linear} embedding goal. To overcome both of these
problems, we propose to compute the weight vectors using a
low-dimensional neighborhood representation. We prove theoretically
that this straightforward and computationally simple modification of
LLE reduces LLE's sensitivity to noise. This modification also
removes the need for regularization when the number of neighbors is
larger than the dimension of the input. We present numerical
examples demonstrating both the perturbation and linear projection
problems, and the improved outputs using the low-dimensional
neighborhood representation.
\end{abstract}


\section{Introduction}
The local linear embedding algorithm (LLE)~\citep{LLE} belongs to a
class of recently developed, non-linear dimension-reducing
algorithms that include Isomap~\citep{ISOMAP}, Laplacian
Eigenmap~\citep{belkin}, Hessian Eigenmap~\citep{HessianEigenMap},
LTSA~\citep{LTSA}, and MVU~\citep{Weinberger}. This group of
algorithms assumes that the data is sitting on, or next to, an
embedded manifold of low dimension within the original
high-dimensional space, and attempts to find an embedding that maps
the input points to the lower-dimensional space. Here a manifold is
defined as a topological space that is locally equivalent to an
Euclidean space. LLE was found to be useful in data
visualization~\citep{LLE,SpeechVisualization} and in image
processing applications, such as image
denoising~\citep{ImageDenoising} and human face
detection~\citep{HumanFaceDetection}. It is also applied in
different fields of science such as chemistry~\citep{Chemistry},
biology~\citep{Biology}, and astrophysics~\citep{Astrophysics}.

LLE attempts to recover the domain structure of the input data set
in three steps. First, LLE assigns neighbors to each input point.
Second, for each input point LLE computes weight vectors that best
linearly reconstruct the input point from its neighbors. Finally,
LLE finds a set of low-dimensional output points that minimize the
sum of reconstruction errors, under some normalization constraints.

In this paper we focus on the computation of the weight vectors in
the second step of LLE. We show that LLE's neighborhood description
captures the structure of the \emph{high}-dimensional space, and not
that of the \emph{low}-dimensional domain. We show two main
consequences of this observation. First, the weight vectors are
highly sensitive to noise. This implies that a small perturbation of
the input may yield an entirely different embedding. Second, we show
that LLE converges to a linear projection of the high-dimensional
input when the number of input points tends to infinity. Numerical
results that demonstrate our claims are provided.

To overcome these problems, we suggest a simple modification to the
second step of LLE, \emph{LLE with low-dimensional neighborhood
representation}. Our approach is based on finding the best
low-dimensional representation for the neighborhood of each point,
and then computing the weights with respect to these low-dimensional
neighborhoods. This proposed modification preserves LLE's principle
of reconstructing each point from its neighbors. It is of the same
computational complexity as LLE and it removes the need to use
regularization when the number of neighbors is greater than the
input dimension.

We prove that the weights computed by LLE with low-dimensional
neighborhood representation are robust against noise. We also prove
that when using the modified LLE on input points sampled from an
isometrically embedded manifold, the pre-image of the input points
achieves a low value of the objective function. Finally, we
demonstrate an improvement in the output of LLE when using the
low-dimensional neighborhood representation for several numerical
examples.

There are other works that suggest improvements for LLE. The
Efficient LLE~\citep{ELLE} and the Robust LLE~\citep{RLLE}
algorithms both address the problem of outliers by preprocessing the
input data. Other versions of LLE, including ISOLLE~\citep{ISOLLE}
and Improved LLE~\citep{ILLE}, suggest different ways to compute the
neighbors of each input point in the first step of LLE. The Modified
LLE algorithm~\citep{MLLE} proposes to improve LLE by using multiple
local weight vectors in LLE's second step, thus characterizing the
high-dimensional neighborhood more accurately.
All of these algorithms attempt to characterize the
\emph{high}-dimensional neighborhoods, and not the
\emph{low}-dimensional neighborhood structure.

Other algorithms can be considered variants of LLE. Laplacian
Eigenmap essentially computes the weight vectors using
regularization with a large regularization constant \citep[see
discussion on the relation between LLE and Laplacian Eigenmap
in][Section~5]{belkin}. Hessian Eigenmap~\citep{HessianEigenMap}
characterizes the local input neighborhoods using the null space of
the local Hessian operator, and minimizes the appropriate function
for the embedding. Closely related is the LTSA
algorithm~\citep{LTSA}, which characterizes each local neighborhood
using its local PCA. These last two algorithms attempt to describe
the low-dimensional neighborhood. However, these algorithms, like
Laplacian Eigenmap, do not use LLE's intuitive approach of
reconstructing each point from its neighbors. Our proposed
modification provides a low-dimensional neighborhood description
while preserving LLE's intuitive approach.

The paper is organized as follows. The description of LLE is
presented in Section~\ref{sec:descriponLLE}. The discussion of the
second step of LLE appears in Section~\ref{sec:highDNeighborhoods}.
The suggested modification of LLE is presented in
Section~\ref{sec:lowd}. Theoretical results regarding LLE with
low-dimensional neighborhood representation appear in
Section~\ref{sec:theory}. In Section~\ref{sec:numericalExamples} we
present numerical examples. The proofs are presented in the
Appendix.

\section{Description of LLE}\label{sec:descriponLLE}
The input data  $X=\{x_1,\ldots,x_N\}, \, x_i\in \R^D$ for LLE is
assumed to be sitting on or next to a $d$-dimensional manifold
$\mathcal{M}$. We refer to $X$ as an $N\times D$ matrix, where each
row stands for an input point. The goal of LLE is to recover the
underlying $d$-dimensional structure of the input data $X$. LLE
attempts to do so in three steps.

First, LLE assigns neighbors to each input point $x_i$. This can be
done, for example, by choosing the input point's $K$-nearest
neighbors based on the Euclidian distances in the high-dimensional
space. Denote by $\{\eta_j\}$ the neighbors of $x_i$. Let the
neighborhood matrix of $x_i$ be denoted by $X_i$, where $X_i$ is the
$K\times D$ matrix with rows $\eta_j-x_i$.

Second, LLE computes weights $w_{ij}$ that best linearly reconstruct
$x_i$ from its neighbors. These weights minimize the reconstruction
error function
\begin{equation}\label{eq:eps}
\varphi_i(w_i)=\|x_i-\sum_j w_{ij} x_j\| ^2\,, 
\end{equation}
where $w_{ij}=0$ if $x_j$ is not a neighbor of $x_i$, and $\sum_j
w_{ij}=1$. With some abuse of notation, we will also refer to $w_i$
as a $K\times 1$ vector, where we omit the entries of $w_i$ for
non-neighbor points. Using this notation, we may write
$\varphi_i(w_i)=w_i'X_i X_i' w_i$.

Finally, given the weights found above, LLE finds a set of
low-dimensional output points $Y=\{y_1,\ldots,y_N\} \in \R^d$ that
minimize the sum of reconstruction errors
\begin{equation}\label{eq:Phi}
\Phi(Y)=\sum_{i=1}^n\|y_i-\sum_j w_{ij} y_j\|^2\,,
\end{equation}
under the normalization constraints $Y'\textbf{1}=0$ and $Y'Y=I$,
where $\textbf{1}$ is vector of ones. These constraints force a
unique minimum of the function $\Phi$.

The function $\Phi(Y)$ can be minimized by finding the $d$-bottom
non-zero eigenvectors of the sparse matrix $(I-W)'(I-W)$, where $W$
is the matrix of weights. Note that the $p$-th coordinate
($p=1,\ldots,d$), found simultaneously for all output points $y_i$,
is equal to the eigenvector with the $p$-smallest non-zero
eigenvalue. This means that the first $p$ coordinates of the LLE
solution in $q$ dimensions, $p<q$, are exactly the LLE solution in
$p$ dimensions~\citep{LLE,thinkGlobally}. Equivalently, if an LLE
output of dimension $q$ exists, then a solution for dimension $p$,
$p<q$, is merely a linear projection of the $q$-dimensional solution
on the first $p$ dimensions.

When the number of neighbors $K$ is greater than the dimension of
the input $D$, each data point can be reconstructed perfectly from
its neighbors, and the local reconstruction weights are no longer
uniquely defined. In this case, regularization is needed and one
needs to minimize
\begin{equation}\label{eq:epsRegularization}
\varphi_i^{\textrm{reg}}(w_i)=\|x_i-\sum_j w_{ij} x_j\| ^2+\delta
\|w_i\|^2\,.
\end{equation}
where $\delta$ is a small constant. \citet{thinkGlobally} suggested
$\delta=\frac{\Delta}{K}\textrm{trace}(X_i X_i')$ with $\Delta\ll
1$. Regularization can be problematic for the following reasons.
When the regularization constant is not small enough, it was shown
by~\citet{MLLE} that the correct weight vectors cannot be well
approximated by the minimizer of $\varphi_i^{\textrm{reg}}(w_i)$.
Moreover, when the regularization constant is relatively high, it
produces weight vectors that tend towards the uniform vectors
$w_i=(1/K,\ldots,1/K)$. Consequently, the solution for LLE with
large regularization constant is close to that of Laplacian
Eigenmap, and does not reflect a solution based on reconstruction
weight vectors \citep[see][Section~5]{belkin}. In addition,
\citet{NLDR} demonstrated that the regularization parameter must be
tuned carefully, since LLE can yield completely different embeddings
for different values of this parameter. However, in real-world data
the dimension of the input is typically greater than the number of
neighbors. Hence, for real-world data, regularization is usually
unnecessary.

\section{Preservation of high-dimensional neighborhood
structure by LLE}\label{sec:highDNeighborhoods}

In this section we focus on the computation of the weight vectors,
which is performed in the second step of LLE. We first show that LLE
characterizes the \emph{high}-dimensional structure of the
neighborhood. We explain how this can lead to the failure of LLE in
finding a meaningful embedding of the input. Two additional
consequences of preservation of the high-dimensional neighborhood
structure are discussed. First, LLE's weight vectors are sensitive
to noise. Second, LLE's output tends toward a linear projection of
the input data when the number of input points tends to infinity.
These claims are demonstrated using numerical examples.

\begin{figure}[!ht]
\vskip 0.2in
\begin{center}
\includegraphics{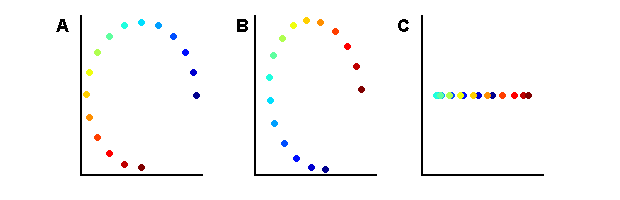}
\caption{The input for LLE is the $16$-point open ring that appears
in~(A). The two-dimensional output of LLE is given in~(B). LLE finds
and preserves the two-dimensional structure of each of the local
neighborhoods. The one-dimensional output of LLE appears in~(C). The
computation was performed using $4$-nearest-neighbors, and
regularization constant $\Delta=10^{-9}$.} \label{fig:ring}
\end{center}
\vskip 0.2in
\end{figure}

We begin by showing that LLE preserves the high-dimensional
neighborhood structure. We use the example that appears in
Fig~\ref{fig:ring}. The input is a sample from an open ring which is
a one-dimensional manifold embedded in $\R^2$. For each point on the
ring, we define its neighborhood using its $4$ nearest neighbors.
Note that its \emph{high}-dimensional ($D=2$) neighborhood structure
is curved, while the \emph{low}-dimensional structure ($d=1$) is a
straight line. The two-dimensional output of LLE (see
Fig.~\ref{fig:ring}) is essentially a reconstruction of the input.
In other words, LLE's weight vectors preserve the curved shape of
each neighborhood.

The one-dimensional output of the open ring is presented in
Fig~\ref{fig:ring}C. Recall that the one-dimensional solution is a
linear projection of the two-dimensional solution, as explained in
section~\ref{sec:descriponLLE}. In the open-ring example, LLE
clearly fails to find an appropriate one-dimensional embedding,
because it preserves the two-dimensional curved neighborhood
structure. We now show that this is also true for additional examples.

\begin{figure}[!ht]
\vskip 0.2in
\begin{center}
\includegraphics{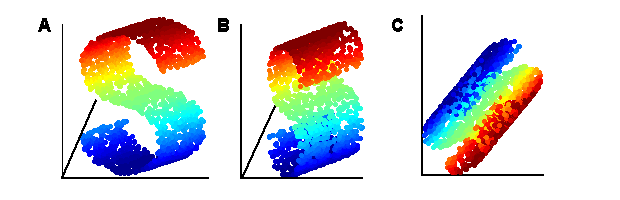}
\caption{(A) LLE's input, a $2000$-point `S' curve. (B) The
three-dimensional output of LLE. It can be seen that LLE finds the
overall three-dimensional structure of the input. (C) The
two-dimensional output of LLE.} \label{fig:scurve}
\end{center}
\vskip 0.2in
\end{figure}

\begin{figure}[!ht]
\vskip 0.2in
\begin{center}
\includegraphics{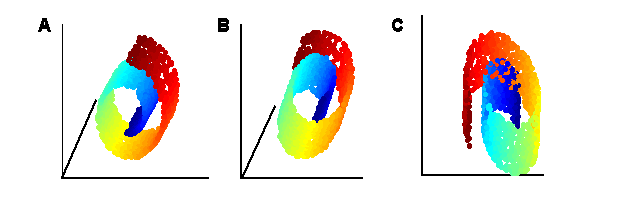}
\caption{(A) LLE's input, a $2000$-point swissroll. (B) The
three-dimensional output of LLE. It can be seen that LLE finds the
overall three-dimensional structure of the input. (C) The
two-dimensional output of LLE.} \label{fig:swissroll}
\end{center}
\vskip 0.2in
\end{figure}

The `S' curve input data appears in Fig~\ref{fig:scurve}A.
Fig~\ref{fig:scurve}B shows that the overall three-dimensional
structure of the `S' curve is preserved in the three-dimensional
embedding. The two-dimensional output of LLE appears in
Fig~\ref{fig:scurve}C. It can be seen that LLE does not succeed in
finding a meaningful embedding in this case. Fig~\ref{fig:swissroll}
presents the swissroll, with similar results.

We performed LLE, here and in all other examples, using the LLE
Matlab code as it appears on the LLE
website~\citep{LLECode}.\footnote{The changes in the Matlab function
\textsl{eigs} were taken into account.} The code that produced the
input data for the `S' curve and the swissroll was also taken from
the LLE website. We used the default values of $2000$-point samples
and $12$-nearest-neighbors. For the regularization constant we used
$\Delta=10^{-9}$. It should be noted that using a large
regularization constant improved the results. However, as discussed
in Section~\ref{sec:descriponLLE}, the weight vectors produced in
this way do not reflect a solution that is based on reconstruction
weight vectors. Instead, the vectors tend toward the uniform vector.

\begin{figure}[!ht]
\vskip 0.2in
\begin{center}
\includegraphics{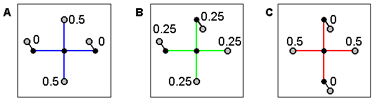}
\caption{The effect of a small perturbation on the weight vector
computed by LLE. All three panels show the same unperturbed
neighborhood, consisting of a point and its four nearest-neighbors
(black points), all sitting in the two-dimensional plane. Each panel
shows a different small perturbation of the original neighborhood
(gray points). All perturbations are in the direction orthogonal to
the plane of the original neighborhood. (A) and (C): Both
perturbations are in the same direction. (B) Perturbations are of
equal size, in opposite directions. The unique weight vector for the
center point is denoted for each case. These three different weight
vectors vary widely, even though the different perturbations can be
arbitrarily small.} \label{fig:pertubationExample}
\end{center}
\vskip 0.2in
\end{figure}

We now discuss the sensitivity of LLE's weight vectors $\{w_i\}$ to
noise. Figure~\ref{fig:pertubationExample} shows that an arbitrarily
small change in the neighborhood can cause a large change in the
weight vectors. This result can be understood by noting how the
vector $w_i$ is obtained. It can be shown~\citep{thinkGlobally} that
$w_i$ equals $(X_i X_i')^{-1}\one$, up to normalization. Sensitivity
to noise is therefore expected when the condition number of $X_i
X_i'$ is large~\citep[see][Section~2]{MatrixComputations}. One way
to solve this problem is to enforce regularization, with its
associated problems (see section~\ref{sec:descriponLLE}). In the
next section we suggest a simple alternative solution to the
sensitivity of LLE to noise.

One more implication of the fact that LLE preserves the
high-dimensional neighborhood structure is that LLE's output tends
to a linear projection of the input data.~\citet{LinearLLE} proved
for a finite data set that when the reconstruction errors are
exactly zero for each of the neighborhoods, and under some
dimensionality constraint, the output of LLE must be a linear
projection of the input data. Here, we present a simple argument
that explains why LLE's output tends to a linear projection when the
number of input points tends to infinity, and show numerical
examples that strengthen this claim. For simplicity, we assume that
the input data is normalized.

Our argument is based on two claims. First, note that LLE's output
for dimension $d$ is a linear-projection of LLE's output for
dimension $D$ (see Section~\ref{sec:descriponLLE}). Second, note
that by definition, the LLE output is a set of points $Y$ that
minimizes the sum of reconstruction errors $\Phi(Y)$. For normalized
input $X$ of dimension $D$, when the number of input points tends to
infinity, each point is well reconstructed by its neighboring
points. Therefore the reconstruction error $\varphi_i(w)$ tends to
zero for each point $x_i$. This means that the input data $X$ tends
to minimize the sum of reconstruction errors $\Phi(Y)$. Hence, the
output points $Y$ of LLE for output of dimension $D$ tend to the
input points (up to a rotation). The result of these two claims is
that any requested solution of dimension $d<D$ tends to a linear
projection of the $D$-dimensional solution, i.e., a linear
projection of the input data.

The result that LLE tends to a linear projection is of asymptotical
nature. However, numerical examples show that this phenomenon can
occur even when the number of points is relatively small. This is
indeed the case for the outputs of LLE shown in
Figs.~\ref{fig:ring}C,~\ref{fig:scurve}C, and~\ref{fig:swissroll}C,
for the open ring, the `S' curve, and the swissroll, respectively.

\section{Low-dimensional neighborhood representation for
LLE}\label{sec:lowd} In this section we suggest a simple
modification of LLE that computes the low-dimensional structure of
the input points' neighborhoods. Our approach is based on finding
the best representation of rank $d$ (in the $l_2$ sense) for the
neighborhood of each point, and then computing the weights with
respect to these $d$-dimensional neighborhoods. In
Sections~\ref{sec:theory} and~\ref{sec:numericalExamples} we show
theoretical results and numerical examples that justify our
suggested modification.

We begin by finding a rank-$d$ representation for each local
neighborhood. Recall that $X_i$ is the $K\times D$ neighborhood
matrix of $x_i$, whose $j$-th row is $\eta_j-x_i$, where $\eta_j$ is
the $j$-th neighbor of $x_i$. We assume that the number of neighbors
$K$ is greater than $d$, since otherwise $x_i$ cannot (in general)
be reconstructed by its neighbors. We say that $X_i^P$ is the best
rank-$d$ representation of $X_i$, if $X_i^P$ minimizes
$\norm{X_i-Y}_2$ over all the $K\times D$ matrices $Y$ of rank $d$.
Let $ULV'$ be the SVD of $X_i$, where $U$ and $V$ are orthogonal
matrices of size $K\times K$ and $D\times D $, respectively, and $L$
is a $K\times D$ matrix, where $L_{jj}=\lambda_j$ are the singular
values of $X_i$ for $j=\min(K,D)$, ordered from the largest to the
lowest, and $L_{ij}=0$ for $i\neq j$. We denote
\begin{equation}\label{eq:ULV}
    U=\left(\begin{array}{cc}U_1, & U_2 \\ \end{array} \right)\,;\;
L=\left(\begin{array}{cc}
  L_1, & 0\\
  0, & L_2
\end{array}\right)
    \,;\;V=\left(\begin{array}{cc}V_1, & V_2 \\ \end{array} \right)
\end{equation}
where $U_1=(u_1,\ldots,u_d)$ and $V_1=(v_1,\ldots,v_d)$ are the
first $d$ columns of $U$ and $V$, respectively, $U_2$ and $V_2$ are
the last $K-d$ and $D-d$ columns of $U$ and $V$ respectively, and
$L_1$ and $L_2$ are of dimension $d\times d$ and $(K-d)\times(D-d)$,
respectively. Then by Corollary 2.3-3 of~\citet{MatrixComputations},
$X_i^P$ can be written as $U_1 L_1 V_1'$.

We now compute the weight vectors for the $d$-dimensional
neighborhood $X_i^P$. By~\eqref{eq:eps}, we need to find $w_i$ that
minimize $w_i' X_i^P {X_i^P} ' w_i$ (see
Section~\ref{sec:descriponLLE}). The solution for this minimization
problem is not unique, since by the construction all the vectors
spanned by $u_{d+1},\ldots,u_K$ zero this function. Thus, our
candidate for the weight vector is the vector in the span of
$u_{d+1},\ldots,u_K$ that has the smallest $l_2$ norm. In other
words, we are looking for
\begin{equation}\label{eq:argmin_w}
 \mathop{\argmin_{w_i\in \textrm{span}\{u_{d+1},\ldots,u_K\}}}_{w_i' \one=1}\hspace{-0.2in} \|w_i\|^2\,.
\end{equation}

Note that we implicitly assume that $\one \notin
\textrm{span}\{u_{1},\ldots,u_d\}$. This is true whenever the
neighborhood points are in \textit{general position}, i.e., no $d +
1$ of them lie in a $(d-1)$-dimensional plane. To understand this,
note that if $\one \in \textrm{span}\{u_{1},\ldots,u_d\}$ then
$(I-\frac{1}{K}\one\one')X_i^P=(I-\frac{1}{K}\one\one')U_1 L_1V_1'$
is of rank $d-1$. Since $(I-\frac{1}{K}\one\one')X_i^P $ is the
projected neighborhood after centering, we obtained that the
dimension of the centered projected neighborhood is of dimension
$d-1$, and not $d$ as assumed, and therefore the points are not in
general position. See also Assumption~(A\ref{as:mu}) in
Section~\ref{sec:theory} and the discussion that follows.

The following Lemma shows how to compute the vector $w_i$ that
minimizes~\eqref{eq:argmin_w}.
\begin{lem}\label{lem:minimzier_w}
Assume that the points of $X_i^P$ are in general position. Then the
vector $w_i$ that minimizes~\eqref{eq:argmin_w} is given by
\begin{equation}\label{eq:w}
w_i=\frac{U_2{U_2 }'\one}{\one'U_2{U_2 }'\one}\,.
\end{equation}
\end{lem}
The proof is based on Lagrange multipliers and appears in
Appendix~\ref{ap:lem}.

Following Lemma~\ref{lem:minimzier_w}, we propose a simple
modification for LLE based on computing the reconstruction vectors
using $d$-dimensional neighborhood representation.

\framebox{
\begin{minipage}{\linewidth}
\textbf{Algorithm: LLE with low-dimensional neighborhood representation}\\ \\

\textbf{Input}: $X$, an $N\times D$ matrix.\\
\textbf{Output}: $Y$, an $N\times d$ matrix.\\
\\ \textbf{Procedure:}\\
\begin{enumerate}
  \item For each point $x_i$ find $K$-nearest-neighbors and compute the neighborhood matrix $X_i$.
  \item For each point $x_i$ compute the weight vector $w_i$ using the $d$-dimensional neighborhood
  representation:
  \begin{itemize}
    \item Use the SVD decomposition to write $X_i=ULV'$.
    \item Write $U_2=(u_{d+1}\,\ldots,u_{K})$.
    \item Compute
    \begin{equation*}w_i=\frac{U_2{U_2 }'\one}{\one'U_2{U_2
    }'\one}\,.\end{equation*}
  \end{itemize}
  \item Compute the $d$-dimension embedding by minimizing $\Phi(Y)$
   (see~\eqref{eq:Phi}).
\end{enumerate}
\end{minipage}
}

Note that the difference between this algorithm and LLE is in step
(2). We compute the low-dimensional neighborhood representation of
each neighborhood and obtain its weight vector, while LLE computes
the weight vector for the original high-dimensional neighborhoods.
One consequence of this approach is that the weight vectors $w_i$
are less sensitive to perturbation, as shown in Theorem~\ref{thm:w}.
Another consequence is that the $d$-dimensional output is no longer
a projection of the embedding in dimension $q, \,q>d$. This is
because the weight vectors $w_i$ are computed differently for
different values of output dimension $d$. In particular, the input
data no longer minimize $\Phi$, and therefore the linear projection
problem does not occur.

From a computational point of view, the cost of this modification is
small. For each point $x_i$, the cost of computing the SVD of the
matrix $X_i$ is $\mathcal{O}(DK^3)$. For $N$ neighborhoods we have
$\mathcal{O}(NDK^3)$ which is of the same scale as LLE for this
step. Since the overall computation of LLE is $\mathcal{O}(N^2 D)$,
the overhead of the modification has little influence on the running
time of the algorithm~\citep[see][Section~4]{thinkGlobally}.

\section{Theoretical results}\label{sec:theory}
In this section we prove two theoretical results regarding the
computation of LLE using the low-dimensional neighborhood
representation. We first show that a small perturbation of the
neighborhood has a small effect on the weight vector. Then we show
that the set of original points in the low-dimensional domain that
are the pre-image of the input points achieve a low value of the
objective function $\Phi$.

We start with some definitions. Let $\Omega\subset \R^d$ be a
compact set and let $f:\Omega\rightarrow \R^D$ be a smooth conformal
mapping. This means that the inner products on the tangent bundle at
each point are preserved up to a scalar $c$ that may change
continuously from point to point. Note that the class of isometric
embeddings is included in the class of conformal embeddings.
Let $\mathcal{M}$ be the $d$-dimensional image of $\Omega$ in
$\R^D$. Assume that the input $X=\{x_1,\ldots,x_N\}$ is a sample
taken from $\mathcal{M}$. For each point $x_i$, define the
neighborhood $X_i$ and its low-dimensional representation $X_i^P$ as
in Section~\ref{sec:lowd}. Let $X_i=ULV'$ and $X_i^P=U_1 L_1 {V_1}'$
be the SVDs of the $i$-th neighborhood and its projection,
respectively. Denote the singular values of $X_i$ by
$\lambda_1^i\geq\ldots\geq\lambda_{K}^i$, where $\lambda_{j}^i=0$ if
$D<j\leq K$. Denote the mean vector of the projected $i$-th
neighborhood by $\mu_i=\frac{1}{K}\one'X_i^P$.

For the proofs of the theorems we require that the local
high-dimensional neighborhoods satisfy the following two
assumptions.
\begin{enumerate}
\renewcommand{\labelenumi}{(A\arabic{enumi})}
  \item For each $i$, $\lambda_{d+1}^i \ll
  \lambda_{d}^i$. \\More specifically, it is enough to demand
$ \lambda_{d+1}^i
<\min\left\{(\lambda_d^i)^2,\frac{\lambda_d^i}{72}\right\}$.
\label{as:lambda}
   \item There is an $\alpha<1$ such that for all $i$,
  $\frac{1}{K}\one'U_1{U_1}'\one<\alpha$.\label{as:mu}
\end{enumerate}
The first assumption states that for each $i$, the neighborhood
$X_i$ is essentially $d$-dimensional. The second assumption was
shown to be equivalent to the requirement that points in each
projected neighborhood are in general position (see discussion in
Section~\ref{sec:highDNeighborhoods}). We now show that this is
equivalent to the requirement that the variance-covariance matrix of
the projected neighborhood is not degenerate. Denote
$S=\frac{1}{K}{X_i^P}'{X_i^P}=\frac{1}{K}V_1 L_1^2 {V_1}'$, then
\begin{equation*}
  \frac{1}{K}\one'U_1{U_1}'\one =\frac{1}{K} \one'(U_1 L_1 {V_1}')(V_1 L_1^{-2} {V_1}') V_1 L_1 {U_1}'\one
   = \mu' S^{-1} \mu \,.
\end{equation*}
Note that since $S-\mu \mu'$ is positive definite, so is $I-
S^{-1/2} \mu \mu' S^{-1/2}$. Since the only eigenvalues of $I-
S^{-1/2} \mu \mu' S^{-1/2}$ are $1$ and $1-\mu' S^{-1} \mu $, we
obtain that $\mu' S^{-1} \mu <1$.

\begin{thm}\label{thm:w}
Let $E_i$ be a $K\times D$ matrix such that $\|E_i\|_F=1$. Let
$\Xt=X_i+\eps E_i$ be a perturbation of the $i$-th neighborhood.
Assume (A\ref{as:lambda}) and~(A\ref{as:mu}) and $\eps<
\min\left(\frac{(\lambda_d^i)^4}{72},\frac{(\lambda_d^i)^2
(1-\alpha)}{72}\right)$ and that $\lambda_1^i<1$. Let $w_i$ and
$\wt$ be the weight vectors for $X_i$ and $\Xt$, respectively, as
defined by~\eqref{eq:argmin_w}. Then
\begin{equation*}
    \norm{w_i-\wt}< \frac{20\eps}{(\lambda_d^i)^2(1-\alpha)}\,.
\end{equation*}
\end{thm}
See proof in Appendix~\ref{ap:Phi}. Note that the assumption that
$\lambda_1^i<1$ can always be fulfilled by rescaling the matrix
$X_i$ since rescaling the input matrix $X$ has no influence on
the value of $w_i$.

\begin{figure}[!ht] \vskip 0.2in
\begin{center}
\includegraphics{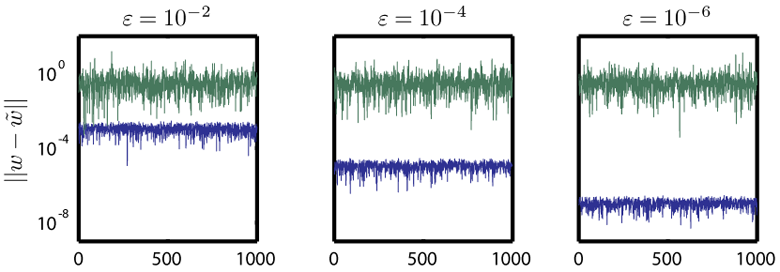}
\caption{The effect of neighborhood perturbation on the weight
vectors of LLE and of LLE with low-dimensional neighborhood
representation. The original neighborhood consists of a point on the
two-dimensional grid and its $4$-nearest neighbors, as in
Fig.~\ref{fig:pertubationExample}. A $4$-dimensional noise matrix
$\eps E$ where $\|E\|_F=1$ was added to the neighborhood for $\eps=
10^{-2},10^{-4}$ and $10^{-6}$, with $1000$ repetitions for each
value of $\eps$. Note that no regularization is needed since $K=D$.
The graphs show the distance between the vector
$w=\left(\frac{1}{4},\frac{1}{4},\frac{1}{4},\frac{1}{4}\right)$ and
the vectors computed by LLE (in green) and by LLE with
low-dimensional neighborhood representation (in blue). Note the log
scale in the $y$ axis. } \label{fig:noise}
\end{center}
\vskip 0.2in
\end{figure}

Fig~\ref{fig:pertubationExample} demonstrates why no bound similar to
Theorem~\ref{thm:w} exists for the weights computed by LLE. In the
example we see a point on the grid with its $4$-nearest neighbors,
where some noise was added. While $\lambda_1\approx\lambda_2\approx
1-\alpha \approx 1$, and $\eps$ is arbitrary, the distance between
each pair of vectors is at least $\frac{1}{2}$. The bound of
Theorem~\ref{thm:w} states that for $\eps=10^{-2},10^{-4}$ and
$10^{-6}$ the upper bounds on the distance when using the
low-dimensional neighborhood representation are is $20\cdot
10^{-2},20\cdot 10^{-4}$ and $20\cdot 10^{-6}$ respectively. The
empirical results shown in Fig~\ref{fig:noise} are even
lower.

For the second theoretical result we require some additional
definitions.

The \emph{minimum radius of curvature} $r_0=r_0(\mathcal{M})$ is
defined as follows:
\begin{equation*}
    \frac{1}{r_0}=\max_{\gamma,t} \left\{ \norm{\ddot{\gamma}(t)}\right\}
\end{equation*}
where $\gamma$ varies over all unit-speed geodesics in $\mathcal{M}$
and $t$ is in a domain of $\gamma$.

The \emph{minimum branch separation} $s_0=s_0(\mathcal{M})$ is
defined as the largest positive number for which
$\norm{x-\tilde{x}}<s_0$ implies $d_\mathcal{M}(x,\tilde{x})\leq \pi
r_0$, where $x,\tilde{x}\in \mathcal{M}$ and
$d_\mathcal{M}(x,\tilde{x})$ is the geodesic distance between $x$
and $\tilde{x}$~\citep[see][for both
definitions]{IsoMapConvergence}.

Define the radius $r(i)$ of neighborhood $i$ to be
\begin{equation*}
r(i)= \max_{j \in \{1,\ldots,K\}} \norm{\eta_j-x_i}
\end{equation*}
where $\eta_j$ is the $j$-th neighbor of $x_i$. Finally, define
$r_{\max}$ to be the maximum over $r(i)$ .

We say that the sample is \emph{dense} with respect to the chosen
neighborhoods if $r_{\max}<s_0$. Note that this condition depends on
the manifold structure, the given sample, and the choice of
neighborhoods. However, for a given compact manifold, if the
distribution that produces the sample is supported throughout the
entire manifold, then this condition is valid with probability increasing
towards $1$ as the size of the sample is increased and the radius of
the neighborhoods is decreased.

\begin{thm}\label{thm:Phi}
Let $\Omega$ be a compact convex set. Let $f:\Omega\rightarrow\R^D$
be a smooth conformal mapping. Let $X$ be an $N$-point sample taken
from $f(\Omega)$, and let $Z=f^{-1}(X)$, i.e.,  $z_i=f^{-1}(x_i)$.
Assume that the sample $X$ is dense with respect to the choice of
neighborhoods and that
assumptions~(A\ref{as:lambda})and~(A\ref{as:mu}) hold. Then, if the
weight vectors are chosen according to~\eqref{eq:w},
\begin{equation}\label{eq:thRfirst}
\frac{\Phi(Z)}{N}= \max_i\lambda_{d+1}^i
\mathcal{O}\left(r_{\max}^2\right)\,.
\end{equation}
\end{thm}
See proof in Appendix~\ref{ap:Phi}.

The theorem states that the original pre-image data $Z$ has a small
value of $\Phi$ and thus is a reasonable embedding, although not
necessarily the minimizer~\citep[see][]{PriceOfNormalization}. This
observation is not trivial from two reasons. First, it is not known
a-priori that $\{f^{-1}(\eta_j)\}$, the pre-image of the neighbors
of $x_i$, are also neighbors of $z_i=f^{-1}(x_i)$. When
short-circuits occur, this need not to be
true~\citep[see][]{IsomapShortCircuits}. Second, the weight vectors
$\{w_i\}$ characterized the projected neighborhood, which is only an
approximation to the true neighborhood. Nevertheless, the theorem
shows the $Z$ has a low $\Phi$ value.

\section{Numerical results}\label{sec:numericalExamples}
In this section we present empirical results for LLE and LLE with
low-dimensional neighborhood representation on some data sets. For
LLE, we used the Matlab code as appears in LLE
website~\citep{LLECode}. The code for LLE with low-dimensional
neighborhood representation is based on the LLE code and differs
only in step (2) of the algorithm and is available in~\url{JR
homepage.}.

\begin{figure}[!ht]
\vskip 0.2in
\begin{center}
\includegraphics{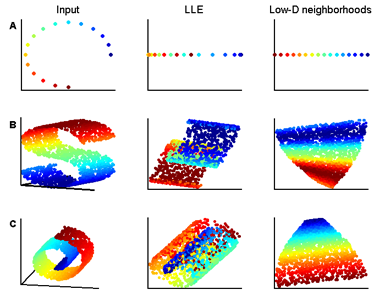}
\caption{The inputs appear in the left column. The results of LLE
appear in the middle column and the the results of LLE with
low-dimensional representation appear in right
column.}\label{fig:numericalAll}
\end{center}
\vskip 0.2in
\end{figure}

We ran LLE with low-dimensional neighborhood representation on the
data sets of the open ring, the `S'-curve and the swissroll that
appear in Figs~\ref{fig:ring}-\ref{fig:swissroll}. We used the same
parameters for both LLE and LLE with low-dimensional neighborhood
representation  ($K=4$ for the open ring and $K=12$ for the
`S'-curve and the swissroll). The results appear in
Fig~\ref{fig:numericalAll}.

\begin{figure}[!ht]
\vskip 0.2in
\begin{center}
\includegraphics{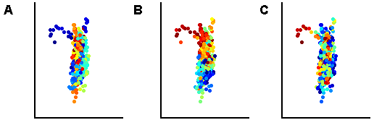}
\caption{The first two dimensions out of the three-dimensional
output of LLE for the faces database appear in all three panels. (A)
is colored according to the right-left pose, (B) is colored
according to the up-down pose, and (C) is colored according to the
lighting direction.}\label{fig:facesLLE}
\end{center}
\vskip 0.2in
\end{figure}

\begin{figure}[!ht]
\vskip 0.2in
\begin{center}
\includegraphics{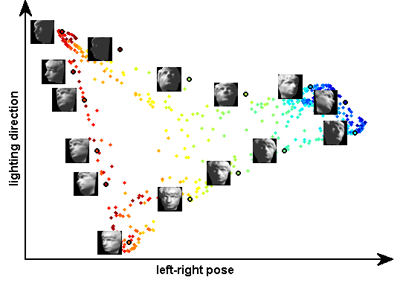}
\caption{The output of LLE with low-dimensional neighborhood
representation is colored according to the left-right pose. LLE with
low-dimensional neighborhood representation also succeeds in finding
the lighting direction. The up-down pose is not fully
recovered.}\label{fig:faces}
\end{center}
\vskip 0.2in
\end{figure}

We ran both LLE and LLE with low-dimensional neighborhood
representation on $64$ by $64$ pixel images of a face, rendered with
different poses and lighting directions. The $698$ images and their
respective poses and lighting directions can be found at the Isomap
webpage~\citep{IsomapHompage}. The results of LLE, with $K=12$, are
given in Fig.~\ref{fig:facesLLE}. We also checked for $K=8,16$; in
all cases LLE does not succeed in retrieving the pose and lighting
directions. The results for LLE with low-dimensional neighborhood
representation, also with $K=12$, appear in Fig~\ref{fig:faces}. The
left-right pose and the lighting directions were discovered by LLE
with low-dimensional neighborhood representation. We also checked
for $K=8,16$; the results are roughly the same.

\acks{This research was supported in part by Israeli Science
Foundation grant. Helpful discussions with Alon Zakai and Jacob
Goldberger are gratefully acknowledged.}

\appendix
\section{Proofs}
\subsection{Proof of Lemma~\ref{lem:minimzier_w}}\label{ap:lem}
\begin{proof}
Write $w_i=\sum_{m=d+1}^K a_m u_m=U_2 a$. The Lagrangian of the
problem can be written as
\begin{equation*}
    L(a,\lambda)= \frac{1}{2} a' {U_2}' U_2 a+\lambda (\one'U_2 a-1)\,.
\end{equation*}
Taking derivatives with respect to both $a$ and $\lambda$, we obtain
\begin{eqnarray*}
  \frac{\partial L}{\partial a} &=& {U_2}' U_2 a-\lambda {U_2 }'\one=a-\lambda {U_2}'\one \,,\\
  \frac{\partial L}{\partial \lambda} &=& \one'U_2
  a-1\,.
\end{eqnarray*}
Hence we obtain that $a=\frac{{U_2 }'\one}{\one'U_2{U_2 }'\one}$.
\end{proof}
\subsection{Proof of Theorem~\ref{thm:w}}
\begin{proof}
The proof of Theorem~\ref{thm:w} consists of two steps. First, we
find a representation of the vector $\wt$, the weight vector of the
perturbed neighborhood, see~\eqref{eq:wtilde}. Then we bound the
distance between $\wt$ and $w_i$, the weight vector of the original
neighborhood.

We start with some notations. For every matrix $A$, let
$\lambda_j(A)$ be the $j$-th singular value of $A$. Note that
$\|A\|_2=\lambda_1(A)$. In this notation, we have
$\lambda_j^i=\lambda_j(X_i)$. Denote by $T={X_i}'X_i$ and
$\Tt={\Xt}'\Xt=T+\eps({X_i}'E_i+{E_i}'X_i)+\eps^2{E_i}'E_i$. Using
the decomposition of~\eqref{eq:ULV}, we may write $T=UL^2U'$ and
$\Tt=\Ut\Lt^2\Ut'$. Note that $\lambda_j(T)=\lambda_j(X_i)^2$. 
Define $U_2$ and $\Ut_2$ to be the $K\times (K-d)$ matrices of the
left-singular vectors corresponding to the lowest singular values,
as in~\eqref{eq:ULV}. 

Note that by assumption, $\lambda_1(E_i)=1$, hence,
$\lambda_1({X_i}'E_i)\leq \lambda_1^i\leq 1$. By Corollary~8.1-3
of~\citet{MatrixComputations},
\begin{equation}\label{eq:difference_in_lambda}
 \lambda_{i}(T)-3\eps \leq  \lambda_{i}(\Tt)\leq \lambda_{i}(T)+3\eps\,.
\end{equation}
Let $\delta=\lambda_d(T)-\lambda_{d+1}(T)-\eps$. By Theorem~8.1-7
of~\cite{MatrixComputations}, there is a $d\times (K-d)$ matrix $Q$
such that $\|Q\|_2\leq \frac{6\eps  }{\delta}$ and such that the
columns of $\widehat{U}_2=(U_2+U_1 Q)(I+Q'Q)^{-1/2}$ are an
orthogonal basis for an invariant subspace of $\Tt$. We want to show
that $\widehat{U}_2$ and $\Ut_2$ spans the same subspaces. To prove
this, we bound the largest singular value of $ \|\widehat{U}_2'\Tt
\widehat{U}_2\|_2$, and the result follows
from~\eqref{eq:difference_in_lambda}.

First, note that
\begin{equation}\label{eq:lambdajIplusQQ}
1-\frac{6\eps}{\delta}<\lambda_j\left((I+Q'Q)^{-1/2}\right)<1+\frac{6\eps}{\delta}\,.
\end{equation}
Hence,
\begin{eqnarray}\label{eq:U2TU2}
   \norm{\widehat{U}_2'\Tt
\widehat{U}_2}_2 &=& \norm{(I+Q'Q)^{-1/2}(U_2+U_1 Q)'\Tt
(U_2+U_1 Q)(I+Q'Q)^{-1/2}}_2 \nonumber\\
  &\leq& \left(1+\frac{6\eps \li}{\delta}\right)^2 \left(\norm{U_2'\Tt U_2}_2+ 2\norm{U_2'\Tt
  U_1Q}_2+\norm{Q'U_1'\Tt  U_1Q}_2\right)\nonumber\\
  &\leq&  \left(1+\frac{6\eps }{\delta}\right)^2\left(
  (\lambda_{d+1}(T)+3\eps )+\frac{\left(6\eps \right)^2}{\delta}+
 \left(\frac{6\eps }{\delta} \right)^2(1+3\eps)
 \right).
\end{eqnarray}

We now obtain some bounds on the size of $\eps$. By assumption we have
$\eps<\frac{(\lambda_d^i)^4}{72} $. Since
assumption~(A\ref{as:lambda}) holds, we may assume that
$\lambda_{d+1}(T)<\frac{\lambda_d(T)}{72}$. Recall that
$\delta=\lambda_d(T)-\lambda_{d+1}(T)-\eps$ and that
$(\lambda_d^i)^2=\lambda_d(T)$. Isolating $\eps$ we obtain that
$\eps<\frac{\lambda_d(T)\delta }{60}$. Similarly, we can show that
$\eps<\frac{\delta^2 }{60}$. We also have $\eps <\frac{\lambda_d(T)
}{72}$, since by assumption $\lambda_d(T)<1$, and similarly,
$\eps<\frac{\delta }{60}$. Summarizing, we have
\begin{equation}
    \eps<\min\left(\frac{\delta}{60}\,,\frac{\lambda_d(T)}{72}\,, \frac{\lambda_d(T)\delta}
    {60}\,,\frac{\delta^2}{60} \right)
\end{equation}

We are now ready to bound the expression in~\eqref{eq:U2TU2}. We
have that $(1+\frac{6\eps}{\delta})< \frac{11}{10}$ since
$\eps<\frac{\delta}{60}$; $ \lambda_{d+1}(T)<
\frac{\lambda_d(T)}{72} $ by assumption; $
3\eps<\frac{\lambda_d(T)}{24}$ since $ \eps <
\frac{\lambda_d(T)}{72}$; $\frac{(6\eps)^2}{\delta} <
\frac{\lambda_d(T)}{120}$ since $\eps<\frac{\delta}{60}$ and also
$\eps<\frac{\lambda_d(T)}{72}$; $\frac{(6\eps)^2}{\delta^2}
<\frac{\lambda_d(T)}{100}$ since $\eps<\frac{\lambda_d(T)\delta
}{60}$ and $\eps< \frac{\delta}{60}$;
$118\frac{\eps^3}{\delta^2}<\frac{\lambda_d(T)}{1000}$ since
$\eps<\frac{\delta}{60}$ and $\eps<\frac{\lambda_d(T)}{72}$.
Combining all these bounds, we obtain
\begin{equation*}
    \norm{\widehat{U}_2'\Tt\widehat{U}_2}_2
    <\frac{\lambda_d(T)}{10}<\lambda_d(T)-3\eps\,.
\end{equation*}
Hence, by~\eqref{eq:difference_in_lambda} we have
$\norm{\widehat{U}_2'\Tt \widehat{U}_2}_2<\lambda_d(\Tt)$. Since
$\widehat{U}_2$ spans a subspace of $K-d$ dimension, it must span
the subspace with the $K-d$ vectors with lowest singular values of
$\Tt$. In other words, $\widehat{U}_2$ spans the same subspace as
$\Ut_2$ or equivalently
$\widehat{U}_2{\widehat{U}_2}'=\Ut_2{\Ut_2}'$. Summarizing, we
obtained that
\begin{equation}\label{eq:wtilde}
    \wt=\frac{\Uh_2{\Uh_2}'\one}{\one'\Uh_2{\Uh_2}'\one}\,.
\end{equation}

We are now ready to bound the  difference between $w_i$ and $\wt$.
\begin{eqnarray*}
  \norm{w_i-\wt}^2 &=& \left\|\frac{U_2{U_2}'\one}{\one' U_2{U_2}'\one}-
  \frac{\Ut_2{\Uh_2}'\one}{\one'\Uh_2{\Uh_2}'\one}\right\|^2  \\
   &=&\frac{1}{\one' U_2{U_2}'\one} -2\frac{\one'U_2 {U_2}' \Uh_2{\Uh_2}'\one}
   {\one' U_2{U_2}'\one \one'
   \Uh_2{\Uh_2}'\one}+\frac{1}{\one'\Uh_2{\Uh_2}'\one}\\
   &=&\frac{\one'(U_2-\Uh_2)(U_2-\Uh_2)'\one}{\one' U_2{U_2}'\one \one'
   \Uh_2{\Uh_2}'\one}
\end{eqnarray*}

We use Assumption~(A\ref{as:mu}) to obtain a bound on $\one'
U_2{U_2}'\one $.  Denote the projection of the normalized vector
$\frac{1}{\sqrt{K}}\one$ on the basis $\{u_j\}$ by
$p_j=\frac{1}{\sqrt{K}}\one' u_i$. We have that
\begin{equation*}
    \|\mu_i\|^2=\frac{1}{K}\norm{\frac{1}{\sqrt{K}}1'U_1 L_1}^2=\frac{1}{K}\sum_{j=1}^d
    \left(p_j\lambda_j^i\right)^2\,.
\end{equation*}
By assumption~(A\ref{as:mu}),
$\|\mu_i\|^2<\frac{\alpha}{K}\left(\lambda_d^i\right)^2$. Hence
$\sum_{j=1}^d p_j^2<\alpha$. Since $\sum_{j=1}^{K}p_j^2=1$, we have
that
\begin{equation}\label{eq:bound1U2U21}
\sum_{j=d+1}^{K} p_j^2=\frac{1}{K}\one' U_2 {U_2}'\one > 1-\alpha\,.
\end{equation}

Similarly, we obtain a bound on $\one' \Uh_2 {\Uh_2}'\one$.
\begin{eqnarray*}
\one' \Uh_2 {\Uh_2}'\one  &\geq& \norm{(I+Q'Q)^{-1/2}U_2'\one}^2-2\left|\one' U_1 Q (I+Q'Q)^{-1}U_2'\one\right| \\
  &\geq&  (1-\frac{6\eps}{\delta})^2 K(1-\alpha)-2K\frac{6\eps}{\delta}(1+\frac{6\eps}{\delta})^2(1-\alpha)^{1/2}\\
   &\geq& \frac{9 K(1-\alpha)}{10}-12K\frac{\eps}{\delta}\left(\frac{11}{10}\right)^2(1-\alpha)^{1/2}\,,
\end{eqnarray*}
where we used $\eps<\frac{\delta}{60}$. Since by assumption
$\eps<\frac{\lambda_d(T)\sqrt{(1-\alpha)}}{72}$, and using the facts
that $\lambda_{d+1}(T)<\frac{\lambda_d(T)}{72}$ and
$\eps<\frac{\lambda_d(T)}{72}$, we obtain that
$\eps<\frac{\delta\sqrt{(1-\alpha)}}{60}$. Hence, $\one' \Uh_2
{\Uh_2}'\one \geq \frac{K(1-\alpha)}{2} $.

Finally, we obtain a bound on $\one'(U_2-\Uh_2)(U_2-\Uh_2)'\one$.
\begin{eqnarray*}
\norm{U_2-\Uh_2}_2 &=& \norm{U_2(I-(I+Q'Q)^{-1/2})+U_1 Q (I+Q'Q)^{-1/2}}_2 \\
   &\leq&    \norm{U_2}_2\norm{ I-(I+Q'Q)^{-1/2}}_2+\norm{U_1}_2\norm{Q}_2\norm{(I+Q'Q)^{-1/2}}_2\\
   &\leq& \frac{6\eps}{\delta}
   +\frac{6\eps}{\delta}(1+\frac{6\eps}{\delta})=\frac{6\eps}{\delta}(2+\frac{6\eps}{\delta})
         \end{eqnarray*}
where the last inequality follows from~\eqref{eq:lambdajIplusQQ},
the fact that for any eigenvector $v$ of $(I+Q'Q)^{-1/2}$ with
eigenvalue $\lambda_v$, $v$ is also eigenvector of
$I-(I+Q'Q)^{-1/2}$ with eigenvalue $1-\lambda_v$, and the fact that
$\|A\|_2=1$ for every matrix $A $ with orthonormal
columns~\citep[see][]{MatrixComputations}. Consequently,
\begin{equation*}
 \norm{(U_2-\Uh_2)'\one}_2\leq
  K\frac{6\eps}{\delta}\left(2+\frac{6\eps}{\delta}\right)<\frac{13K\eps}{\delta}\,
\end{equation*}
where we used $\eps<\frac{\delta}{60}$.

Combining these results, we have
\begin{equation*}
    \norm{w_i-\wt}< \frac{(13K\eps)/\delta}
    {(K(1-\alpha))/\sqrt{2}}<\frac{20\eps}{\lambda_d(T)(1-\alpha)}\,,
\end{equation*}
where we used $\frac{21}{20\lambda_d(T)}>\frac{1}{\delta}$.

\end{proof}

\subsection{Proof of Theorem~\ref{thm:Phi}}\label{ap:Phi}
\begin{proof}
Since $\Phi(Z)=\sum_{i=1}^N\norm{ \sum_{j} w_{ij} (z_j-z_i)}^2 $, we
bound each summand separately in order to obtain a global bound.

Let the induced neighbors of $z_i=f^{-1}(x_i)$ be defined by
$(\tau_1,\ldots,\tau_K)=(f^{-1}(\eta_1),\ldots,f^{-1}(\eta_K))$.
Note that a-priori, it is not clear that $\tau_j$ are neighbors of
$z_i$. Let $J$ be the Jacobian of the function $f$ at $z_i$. Since
$f$ is a conformal mapping, $J'J=c(z_i)I$, for some positive
$c:\Omega\rightarrow \R$. Using first order approximation we have
that $\eta_j-x_i =
J(\tau_j-z_i)+\mathcal{O}\left(\norm{\tau_j-z_i}^2\right)$. Hence,
for $w_i$ we have,
\begin{equation*}
  \sum_{j=1}^K w_{ij} (\tau_j-z_i) =
  \sum_{j=1}^K w_{ij} J'(\eta_j-x_i) +\mathcal{O}\left(\max_j\norm{\tau_j-z_i}^2\right)   \,.
\end{equation*}
Thus we have
\begin{equation}\label{eq:normTauMinusZi}
 \norm{ \sum_{j=1}^K w_{ij} (\tau_j-z_i)}^2 =
  \norm{\sum_{j=1}^K w_{ij} J'(\eta_j-x_i)}^2 +
  \norm{\sum_{j=1}^K w_{ij} J'(\eta_j-x_i)}\mathcal{O}\left(\max_j\norm{\tau_j-z_i}^2\right)
  \,.
\end{equation}

We bound $\norm{\sum_{j=1}^K w_{ij} J'(\eta_j-x_i)} $ for the vector
$w_i$ that minimizes~\eqref{eq:argmin_w}. Note that
by~\eqref{eq:ULV}, $\sum_{j=1}^K w_{ij} J'(\eta_j-x_i) = w_i'X_i^P
J+w_i' U_2 L_2 V_2' J$. However, by construction $w_i'X_i^P=0$.
Hence
\begin{equation*}
 \norm{\sum_{j=1}^K w_{ij} J'(\eta_j-x_i)} = \norm{w_i' U_2
L_2 V_2' J} \leq \norm{w_i}\norm{U_2 L_2 {V_2}' J}_2 \leq
\frac{\norm{w_i} \lambda_{d+1}^i}{\sqrt{c(z_i)}}\,,
\end{equation*}
where we used the facts that $\|Ax\|_2\leq \|A\|_2\|x\|_2$ for a any
matrix $A$, and that $\|A\|_2=1$ for a matrix $A$ with orthonormal
columns~\citep[see Section~2 of][for both]{MatrixComputations}.
Substituting in~\eqref{eq:normTauMinusZi}, we obtain
\begin{equation*}
    \norm{ \sum_{j=1}^K w_{ij} (\tau_j-z_i)}^2 \leq \frac{\norm{w_i}^2
    (\lambda_{d+1}^i)^2}{c(z_i)}+\norm{w_i}
    \lambda_{d+1}^i\mathcal{O}\left(\max_j\norm{\tau_j-z_i}^2\right)\,.
\end{equation*}
Since assumption~(A\ref{as:mu}) hold, it follows
from~\eqref{eq:bound1U2U21} that $\norm{w_i}^2=\frac{1}{\one' U_2
{U_2}'\one}< \frac{1}{K(1-\alpha)} $.

As $f$ is an conformal mapping, we have $ c_{\min}\norm{\tau_j-z_i}
\leq d_{\mathcal{M}}(\eta_j,x_i)$, where $d_{\mathcal{M}}$ is the
geodesic metric and $c_{\min}>0$ is the minimum of the scale
function $c(z)$ that measures the scaling change of $f$ at $z$ . The
minimum $c_{\min}$ is attained as $\Omega$ is compact. The last
inequality holds true since the geodesic distance
$d_{\mathcal{M}}(\eta_j,x_i)$ is equal to the integral over $c(z)$
for some path between $\tau_j$ and $z_i$.

The sample is assumed to be dense, hence $\norm{\tau_j-x_i} < s_0$,
where $s_0$ is the \emph{minimum branch separation} (see
Section~\ref{sec:theory}). Using \citet{IsoMapConvergence}, Lemma~3,
we conclude that
\begin{equation*}
\norm{\tau_j-z_i}\leq \frac{1}{c_{\min}}d_{\mathcal{M}}(\eta_j,x_i)
<\frac{\pi}{2c_{\min}}\norm{\eta_j-x_i}\,.
\end{equation*}

Since assumption~(A\ref{as:lambda}) holds, and
\begin{equation*}
    r(i)^2=\max_j \|\eta_j-x_i\|^2
    \geq\frac{1}{K}\sum_{j=1}^K\|\eta_j-x_i\|^2=\|X_i\|_F^2=\frac{1}{K}\sum_{j=1}^{K}(\lambda_j^i)^2\geq
    \frac{d}{K}(\lambda_d^i)^2\,,
\end{equation*}
we have $\lambda_{d+1}\ll r(i)$. Hence $\norm{ \sum_{j=1}^K w_{ij}
(\tau_j-z_i)}^2= \lambda_{d+1}^i
    \mathcal{O}\left(r(i)^2\right)$.
\end{proof}


\vskip 0.2in


\end{document}